%% file: ijcai24.tex
\documentclass{article}
\usepackage{ijcai24}

\usepackage{times}
\usepackage{soul}
\usepackage{url}
\usepackage[hidelinks]{hyperref}
\usepackage[utf8]{inputenc}
\usepackage[small]{caption}
\usepackage{graphicx}
\usepackage{amsmath}
\usepackage{amsthm}
\usepackage{booktabs}
\usepackage{algorithm}
\usepackage{algorithmic}
\usepackage{color}

\usepackage[utf8]{inputenc}
\usepackage{tikz}
\usepackage{forest}
\usetikzlibrary{trees,positioning,shapes,shadows,arrows.meta}

\title{A Survey on Large Language Model Hallucination via a Creativity Perspective}

\author{
Xuhui Jiang$^{1,2,3}$
\and
Yuxing Tian$^3$
Fengrui Hua$^3$
Chengjin Xu$^3$
Yuanzhuo Wang$^1$
Jian Guo$^3$\\
\affiliations
$^1$CAS Key Laboratory of AI Safety \& Security, Institute of Computing Technology, CAS\\
$^2$School of Computer Science and Technology, University of Chinese Academy of Science\\
$^3$International Digital Economy Academy, IDEA Research\\
\emails
\{jiangxuhui19g, wangyuanzhuo\}@ict.ac.cn,
\{tianyuxing, huafengrui, xuchengjin, guojian\}@idea.edu.cn
}

\begin{document}

\maketitle

\begin{abstract}
Hallucinations in large language models (LLMs) are always seen as limitations. However, could they also be a source of creativity? This survey explores this possibility, suggesting that hallucinations may contribute to LLM application by fostering creativity. This survey begins with a review of the taxonomy of hallucinations and their negative impact on LLM reliability in critical applications. Then, through historical examples and recent relevant theories, the survey explores the potential creative benefits of hallucinations in LLMs. To elucidate the value and evaluation criteria of this connection, we delve into the definitions and assessment methods of creativity. Following the framework of divergent and convergent thinking phases, the survey systematically reviews the literature on transforming and harnessing hallucinations for creativity in LLMs. Finally, the survey discusses future research directions, emphasizing the need to further explore and refine the application of hallucinations in creative processes within LLMs.
\end{abstract}

\input{sec1}
\input{sec2}

\input{sec3}

\input{sec4}

\input{sec5}

\input{sec6}
\bibliographystyle{named}
\bibliography{ijcai24}

\end{document}

%% file: sec1.tex

\section{Introduction}
Recent advancements in artificial intelligence have propelled large language models (LLMs) into the spotlight, marking a significant leap in their ability to comprehend and generate natural language. This surge in progress has sparked a global wave of research and practical applications, highlighting the transformative impact of LLMs across various domains. Among these advancements, the phenomenon of hallucination within LLMs has emerged as a focal point of investigation. Characterized by the models' tendency to produce unfounded or misleading information without solid data backing, hallucination poses a challenge to the reliability and applicability of LLMs~\cite{ye2023cognitive}. Despite the increasing volume of research, the comprehensive understanding of hallucination for LLMs remains to be explored.

This study begins by outlining the taxonomies of hallucinations in LLMs, setting the stage for an in-depth review of key research efforts aimed at identifying and reducing these occurrences. It highlights how existing literature predominantly views hallucinations as detrimental, advocating for strategies to minimize their presence, particularly in serious application scenarios like legal and financial.

However, a key question raises and provokes deep reflection: “\textit{Is hallucination in LLMs always harmful, or does creativity hide in hallucinations?}” Different from previous surveys or studies about hallucination, this paper revisits the phenomenon from a positive perspective. In addition to the negative impacts of hallucination on the reliability of LLMs, this paper recognizes a trend in research on the creativity of LLMs and explores the interplay between hallucination and creativity, as well as how to unearth the value of LLM hallucination from the perspective of creativity.

In our exploration of the interplay between LLMs' hallucinations and creativity, we scrutinize notable historical examples where hallucinations have catalyzed creative breakthroughs. By examining these instances, we aim to uncover the complex dynamics between human creativity and hallucination, drawing insights from cognitive science underpinned by pertinent scholarly work. Furthermore, this paper reviews recent studies that focus on this specific interplay in the realm of LLMs, underscoring this critical interplay. This analysis lays the groundwork for a deeper comprehension of the symbiotic relationship between hallucination and creativity.

Recognizing this interplay, this paper delves into evaluating its significance and value. We begin by examining the concept and evaluation of creativity through the lens of cognitive science, followed by a thorough review of the studies on LLM creativity, to understand its critical role and potential. This discussion leads to the assertion: "\textit{While minimizing hallucination risks, it's vital to assess and harness its creative potential, maximizing the value of LLM hallucinations.}"

To achieve the transformation from hallucination to creativity, this paper follows the divergent and convergent phases of cognitive science to systematically review the literature on harnessing hallucination in LLMs for creativity. Specifically, we thoroughly explore the divergent phase, focusing on the methods and research that enhance LLMs' ability to generate creative hallucinations. Subsequently, we delve into the convergent phase, examining how these hallucinations can be critically assessed and refined into valuable creative outputs. Through this bifocal approach, the study highlights practical implications and potential future directions in leveraging LLM hallucinations for creative endeavors.

In conclusion, this paper casts an eye toward the horizon of future research, contemplating strategies to broaden and deepen the exploration of the interplay between hallucination and creativity within LLMs.

\begin{figure}[t]
	\centering
	\includegraphics[width=1\columnwidth]{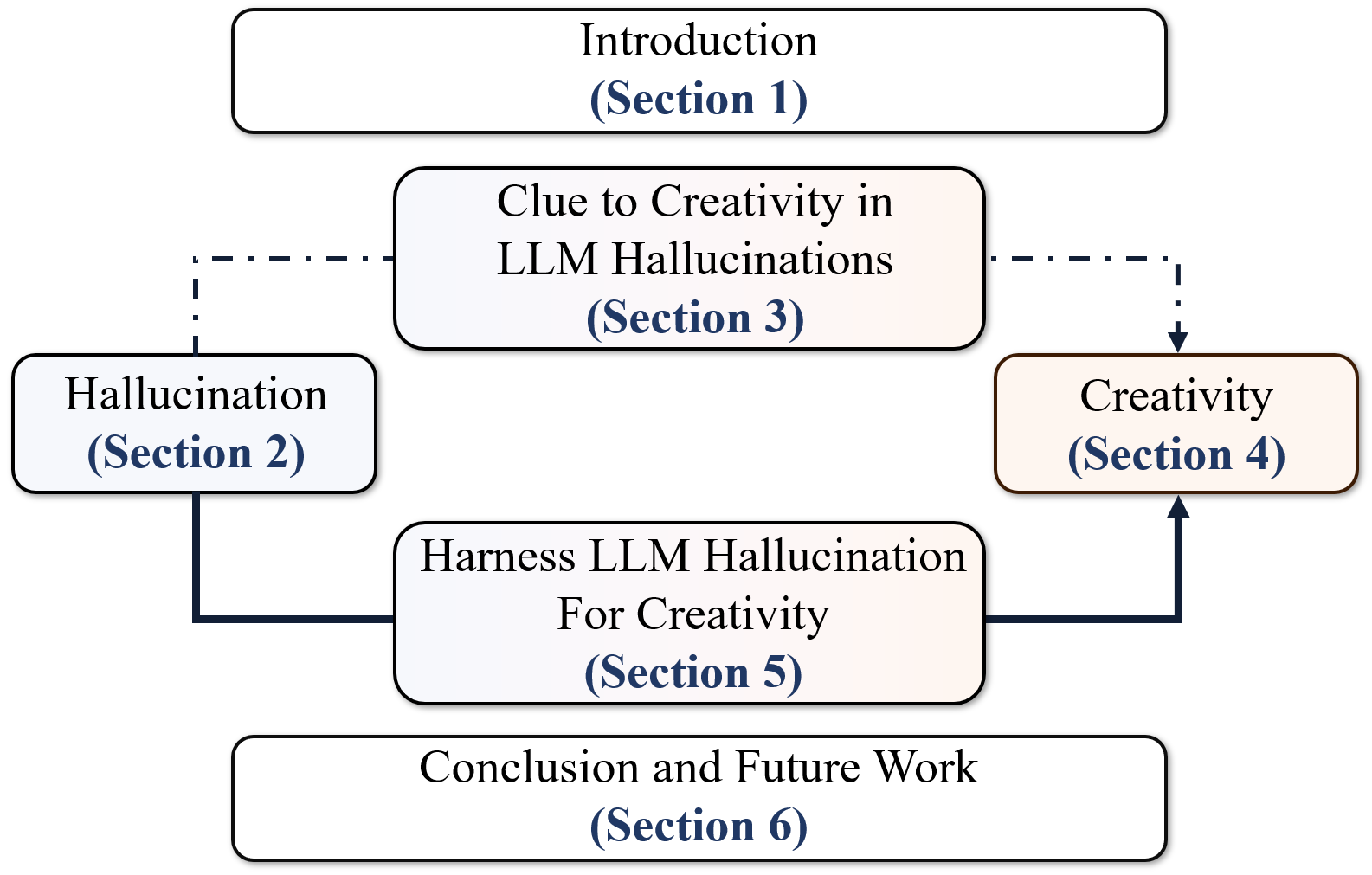}
	\caption{General framework of the survey.}
	\label{FIG:FRAMEWORK}
         \vspace{-10pt}
\end{figure}

The rest of this paper is organized as shown in Figure~\ref{FIG:FRAMEWORK}. Section~\ref{sec2} reviews existing LLM hallucination research. Section~\ref{sec3} questions the negativity of hallucinations and explores their interplay with creativity. Section~\ref{sec4} covers the definition and assessment of creativity in LLMs. Section~\ref{sec5} looks at harnessing the interplay between hallucination and creativity. Finally, Section~\ref{sec6} outlines future directions for this field.

%% file: sec2.tex
\section{Hallucination in the Era of LLM}~\label{sec2}
The advent of LLMs like ChatGPT and LLaMA \cite{touvron2023llama} marks a significant milestone in the field of AI. These models, trained on vast datasets, have demonstrated remarkable proficiency in generating coherent text and opening new frontiers in AI-assisted writing, conversation, and content creation. Their versatility and efficiency have made them indispensable in both academic research and commercial applications.  However, the ascendancy of LLMs brings with it a complex challenge: the phenomenon of 'hallucination', where models confidently generate incorrect or irrelevant information. This often stems from issues like erroneous information in the training data or the problematic alignment process \cite{zhang2023siren}. As LLMs are increasingly applied in various aspects of our lives, comprehensively understanding, detecting, and mitigating hallucinations has become crucial, serving not only as a focal point in academic research but also as a key prerequisite for the reliable deployment of these powerful AI systems.

\subsection{Hallucination Taxonomy}
In dissecting the phenomenon of hallucinations within LLMs, it is crucial to categorize these instances for targeted and effective resolutions. A widely accepted classification method \cite{ye2023cognitive} identifies two main types of hallucinations: factuality hallucinations and faithfulness hallucinations. Factuality hallucinations encompass instances where the content generated by LLMs diverges from verifiable real-world facts. This category manifests in two primary forms. The first is factual inconsistency, where the model's output directly contradicts known real-world information. The second form is factual fabrication, characterized by the generation of content that has no basis in reality and cannot be substantiated with factual data. On the other hand, faithfulness hallucinations occur when the content produced by LLMs does not align with the user's instructions or the context they have previously generated. This category includes three distinct phenomena. Instruction inconsistency is observed when the model's output strays from the user's specific directives. Context inconsistency arises when the output, although possibly factually correct, is irrelevant or inappropriate for the given situation. Lastly, logical inconsistency refers to flaws in the model's reasoning processes or a mismatch between the reasoning and the final output. Similarly, the work from \cite{ji2023survey,dziri2021neural} also echoes these perspectives, categorizing them as intrinsic hallucinations and extrinsic hallucinations. Additionally, \cite{zhang2023siren} categorizes hallucinations according to their types of conflict: input-conflicting, where responses go against user inputs; context-conflicting, involving contradictions within the generated context; and fact-conflicting, where outputs clash with established factual knowledge.

Furthermore, several studies have categorized hallucinations across different applications involving LLMs. \cite{ye2023cognitive} have provided a summary of representative hallucinations in numerous downstream tasks, such as machine translation, question and answer, dialog systems, and summarization systems, among others. The study by \cite{rawte2023survey} provides a structured summary and discussion, categorizing and addressing hallucinations specifically within LLMs, multilingual LLMs, and domain-specific LLMs. These classifications provide a clear framework for understanding the hallucinations that occur in LLMs during content generation and reveal the challenges and limitations faced by LLMs in various tasks.

\subsection{Hallucination Detection}~\label{hallucination_detection}
Catering to the phenomenon of hallucination, existing studies have proposed various detection and evaluation strategies. Reflecting the proposed classification of hallucinations, \cite{huang2023survey} have also summarized the corresponding detection methods for each category. For factuality hallucinations, they recommend strategies that involve retrieving external facts and implementing uncertainty estimation. As for faithfulness hallucinations, they also delineate several approaches, including classifier-based, QA-based, and prompting-based metrics, to sidestep the potential pitfalls of contradictory outputs. \cite{zhang2023siren} categorizes the approaches into two types: generation and discrimination. Generation methods consider hallucination as a generation characteristic, and evaluate the generated texts from LLMs, while discrimination approaches assess LLMs' capability to distinguish between truthful and hallucinated statements. Additionally, research from \cite{ye2023cognitive} proposes methods like reasoning classifiers, uncertainty measures, and self-assessment for identifying hallucinations. 

Furthermore, some research efforts focus on creating benchmarks specifically designed for various application scenarios of LLMs in hallucination detection. HaluEval\cite{li2023halueval} presents a large collection of generated and human-annotated hallucinated samples for assessing LLMs' hallucination detection capabilities, revealing that incorporating external knowledge or adding reasoning steps can improve recognition accuracy. TruthfulQA \cite{lin2021truthfulqa}extends its assessment to 38 domains including law, finance, and politics, employing both manual methods and automated fine-tuning of LLMs for evaluation. Considering the serious consequences in healthcare applications, Med-HALT \cite{umapathi2023med} proposes a two-tiered approach, encompassing Reasoning Hallucination Tests (RHTs) and Memory Hallucination Tests (MHTs), to evaluate the presence of hallucinations. In the scenarios mentioned, such as law, finance, and healthcare, hallucinations are typically detrimental. However, it's worth contemplating whether, in certain domains, hallucinations might not always need to be viewed as problematic.

\subsection{Hallucination Reduction}~\label{hallucination_reduction}
Drawing from the detailed analysis of hallucination types and detection methods in LLMs outlined in the preceding sections, some research has focused on developing targeted strategies to reduce these hallucinations. The work by \cite{zhang2023siren}  classifies approaches to mitigate hallucinations in LLMs based on the timing of their application within the LLM life cycle. For instance, during the reinforcement learning from human feedback (RLHF) stage, a specific reward score targeting hallucinations can be designed and directly optimized through reinforcement learning. In the model inference stage, decoding strategies and the retrieval of external knowledge can be employed for retrieval-based enhancement. \cite{ye2023cognitive} summarizes five methods for addressing hallucinations, including parameter adaptation and leveraging external knowledge, and apply these approaches across various downstream tasks. \cite{tonmoy2024comprehensive} presents an elaborate taxonomy based on a range of parameters, including dataset utilization, common tasks, feedback mechanisms, and types of retrievers. This systematic classification effectively differentiates between the various specialized approaches developed for mitigating hallucination issues in LLMs. It also provides a detailed summary of various strategies based on prompt engineering for reducing hallucinations, including retrieval augmented generation (RAG), and self-refinement.

Particularly, some research has focused on exploring how knowledge graphs (KGs) can aid LLMs in reducing hallucinations\cite{sun2023think} \cite{shi2023hallucination}. \cite{agrawal2023can} argues that KG-augmented retrieval techniques effectively address hallucination issues by expanding the model's knowledge with non-parametric factual knowledge. These studies not only enhance our understanding of the operational mechanisms of LLMs but also provide critical guidance for improving the accuracy and reliability of models in practical applications.


%% file: sec3.tex
\section{The Creativity Hidden in Hallucination}~\label{sec3}
While existing studies have predominantly focused on identifying and mitigating hallucinations in LLMs, often perceiving these as drawbacks, a crucial question arises: \textit{Is hallucination in LLMs always harmful, or does creativity hide in hallucination?}
This paper explores this vital question, aiming to understand the hallucinations of LLMs and harness their potential creative value by reviewing historical cases, cognitive science literature, and recent advancements in LLM research.

\begin{figure}[t]
	\centering
	\includegraphics[width=1\columnwidth]{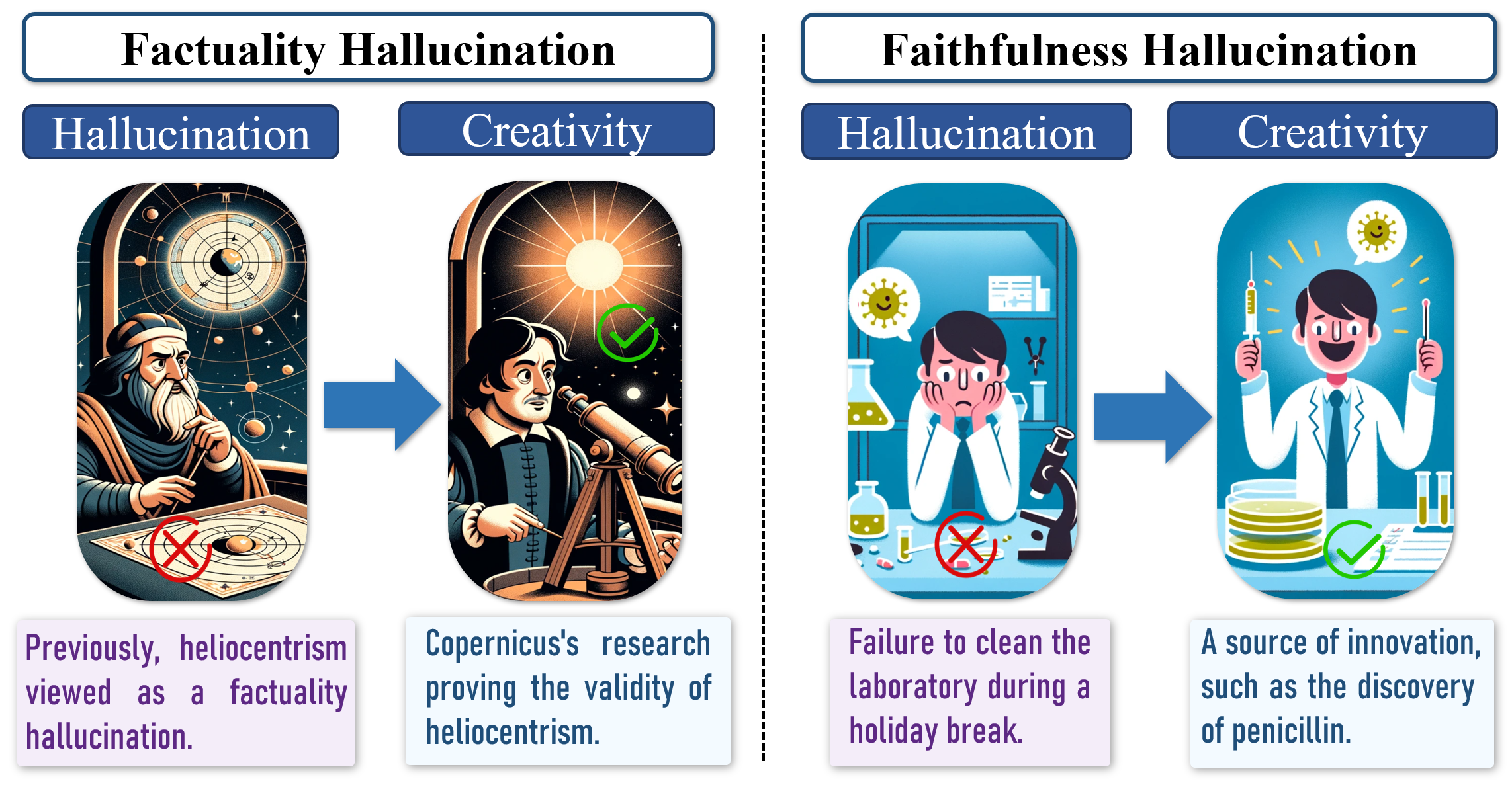}
	\caption{Cases of relations between hallucination and creativity.}
	\label{FIG:CASE}
        \vspace{-10pt}
\end{figure}

Historical examples, in which diverse forms of hallucinations have sparked revolutionary discoveries, offer insightful parallels and serve as a guide in understanding the potential creative value of hallucinations in LLMs.
For instance, as shown in Figure~\ref{FIG:CASE}, consider the notion of factuality hallucinations.
Historically, the shift from a geocentric to a heliocentric model of the solar system was a monumental change in scientific thought. Initially, heliocentrism was dismissed as a factual error, much like how LLMs might generate seemingly erroneous information. However, just as Copernicus's heliocentric model eventually revolutionized astronomy, hallucinations can lead to novel ideas, challenging conventional wisdom.
Similarly, faithfulness hallucinations in LLMs can be likened to accidental discovery. For example, Alexander Fleming's unintended experiment resulted in a groundbreaking medical breakthrough: penicillin.
These examples highlight how both factuality and faithfulness hallucinations could be pivotal in driving creativity.

Expanding upon these historical insights, the clue of the interplay between hallucination and creativity can also be explored in cognitive science studies.
The hallucinations experienced by humans are seen as crucial for simulating creative thought processes. For example, research~\cite{benedek2014create} on human creativity indicates that creative thinking involves both the activation of the left prefrontal cortex, known for its role in imaginative thinking, and the engagement of the right hippocampus, essential for memory processing.
Such insights underscore the idea that creativity is not merely about retrieving information but recombining and expanding upon existing knowledge, which is similar to hallucination~\cite{ye2023cognitive}.
Furthermore, neuroscience research on creativity~\cite{music}, exemplified by studies of improvisation in jazz musicians, parallels this concept. By monitoring different brain regions, researchers have discovered that improvisation activates the same areas as pseudo-random motor movements. This similarity suggests a link between the spontaneity of pseudo-random movements and creative processes, further implying an interplay between hallucinations and creativity. These cognitive science findings also provide valuable insights for research in LLMs.

In the realm of LLMs, there's an increasing focus on the interplay between hallucinations and creativity. As Sam Altman, CEO of OpenAI, elucidates in ~\cite{AltmanSalesforce2023}, there is a profound and often overlooked connection between the hallucinatory phenomena in LLMs and their creative output. Altman argues that LLM hallucination is a cornerstone in unleashing the creative potential of AI.
Recent studies also support this view. The research~\cite{lee2023mathematical} provides a theoretical reference for understanding the interconnectedness of hallucination and creativity in LLMs, substantiating their correlation through rigorous mathematical analysis.
Two surveys also support this point, the research~\cite{wang2024lighthouse} suggests that this interplay is beneficial, while~\cite{rawte2023survey} portrays hallucination models as 'collaborative creative partners'.
Even though their outputs might deviate from factual or instruction, they serve as catalysts for innovative thought.

Despite the emerging evidence linking hallucination and creativity in LLMs, research in this field is still limited. There is a significant need for further research to both deepen our understanding of the value of the link between hallucination and creativity in LLMs and to explore how this interplay can be effectively harnessed for innovative applications.


%% file: sec4.tex
\section{The Creativity of Large Language Models}~\label{sec4}
Regarding the interplay between hallucinations and creativity in LLM, establishing a definitive concept of creativity becomes essential. However, creativity has long been a challenging concept to both define and quantify due to its multifaceted and intricate nature. In this section, we will explore definitions of creativity as understood in cognitive science, examine the approaches employed to measure creativity, and highlight recent studies that investigate LLM creativity. 


\subsection{Definition of Creativity}~\label{sec:definition}
While \cite{Treffinger} presents an extensive array of over 100 different definitions of creativity, it is noteworthy that most studies in this field typically utilize only a select few of these definitions.
In cognitive science, creativity is often conceptualized from four distinct perspectives: cognitive processes associated with creativity (later in this paper referred to as ‘process’), personal characteristics of creative individuals (‘person’), creative products or outcomes (‘product’) and the interaction between the creative individual and the context or environment (‘press’)~\cite{Couger1993UnStructuredCI}. In this paper, we mainly introduce the 'process' and 'product', as they are particularly pertinent to understanding the creativity of LLM, while the 'person' and 'press' categories are more aligned with human creativity.

Regarding the process perspective of creativity, \cite{Torrance1977CreativityIT} defines it as the process of identifying problems or gaps in knowledge, developing hypotheses or propositions, testing and validating hypotheses, and ultimately sharing the findings. Similarly, \cite{RAT} suggests that creativity involves combining associative elements into novel configurations that fulfill the requirements of a given task. \cite{Guilford} views creativity as a type of problem-solving and differentiates between two kinds of cognitive operations: divergent and convergent production. Divergent production is characterized by a wide-ranging search for multiple logical solutions or alternatives to open-ended problems, while convergent production involves a narrow search for a single, precise solution to a problem where one specific answer is needed. They posit that divergent production processes are more closely associated with effective creative thinking.

As the definition of creativity towards the product, \cite{Khatena} defines creativity as the construction or organization of ideas, thoughts, and feelings into unusual and associative connections through the power of imagination. \cite{Gardner} posits that creative individuals possess the capacity to solve problems, fashion products, or formulate new questions in ways that are novel yet acceptable within a specific cultural context. Creativity is also perceived as the ability to generate or conceive something original and adaptive to task constraints, while also being of high quality, useful, aesthetically pleasing, and novel.

Researchers often distinguish between two main types of creativity: everyday creativity, or "little-C," common to nearly everyone, and eminent creativity, or "big-C," which is seen in significant historical figures. Expanding on this, the Four C model\cite{fourC} of creativity introduces "mini-c," representing the creative learning process, and "Pro-c," indicating professional-level creative expertise.

Despite the varied perspectives in defining creativity, a consensus exists among researchers regarding its core attributes. It is widely acknowledged that creativity is characterized by the generation of responses that are both novel and useful. 
However, the precise interpretation of these terms remains a subject of ongoing discourse.

\subsection{Approaches for measuring creativity}
In the field of cognitive science, measuring creativity is proven to be a challenging task. These challenges stem from the subjectivity of creativity and the variety of environments in which it is manifested. Based on the various definitions of creativity, researchers have developed many different methods to measure it. These methods are divided into four categories, representing the four main dimensions of the creativity definition: process, product, person, and press. Here, we focus on the widely adopted process and product approaches.

The process approaches in creativity measurement focus on the 
specific cognitive processes and structures that are conducive to creative production. Divergent thinking tests (DAT)~\cite{DAT} have been most widely used for measuring creative processes or creativity-relevant skills. Examples of these tests include the Associative chain test (ACT)~\cite{ACT}, the Torrance Tests of Creative Thinking (TTCT)\cite{Torrance1977CreativityIT}, the Structure of the Intellect Divergent Production Tests (SOI)~\cite{Guilford}, Wallach-Kogan Creativity Tests and the Creativity Assessment Packet (CAP). These tests, which are also known as measures of ideational fluency, include open or ill-structured problems that require individuals to generate as many responses as possible, which are then scored to capture fluency (number of responses), originality (statistical rarity), flexibility (number of different categories) and elaboration (amount of detail). Hence, the focal point of divergent thinking tests is not only to consider the amount of responses but also the quality of these responses.
Besides, there are some convergence thinking tasks that consider creative ideas are achieved by forming mutually remote associative elements into new and useful combinations, like Remote Associates Test (RAT)~\cite{RAT}, Bridge-the-Associative-Gap Task (BAG)~\cite{BAG}. However, the fundamental psychometric assumptions and underlying cognitive processes involved in them remain controversial.


The product approaches are supported by the rationale that a comprehensive assessment of an individual’s creative capabilities should include the measurement of their concrete creations. Central to this method is the Consensual Assessment Technique (CAT)~\cite{CAT} and widely implemented in studies focusing on the products of creativity. Distinct from other methods of assessment, the CAT does not rely on predetermined theoretical frameworks of creativity. Instead, it utilizes the informed judgment of experts to evaluate the creative outputs within their relevant fields. This unique characteristic of the CAT allows it to incorporate a range of perspectives.
While the CAT for measuring creativity is known for its high inter-rater reliability, several situational variables like the number of tasks and the performance domain can impact this reliability. Furthermore, the practicality of implementing the CAT faces numerous challenges. The selection and expertise level of judges are subjects of ongoing debate, with the consensus among judges notably swayed by their expertise. Additionally, judges personalities, as evidenced in~\cite{John}, can influence creativity ratings, with more agreeable judges tending towards leniency. Cultural disparities among judges also play a role, as~\cite{Niu2001CULTURALIO} found in their comparison of American and Chinese judges' evaluations of artistic creativity. Judges may also display biases, particularly when evaluating their own work.


\subsection{Recent works on evaluating LLM's creativity}~\label{creativity_evaluate}
The section mentioned above delves into the approaches to measure human creativity from a cognitive science perspective. However, since there are differences between LLM and humans, it might lead to irrelevant responses or serious logical issues, requiring us to additionally assess these approaches. This predicament raises a question: \textit{how can we adapt these measures to effectively evaluate the creativity of LLM?} In light of LLM's burgeoning potential, researchers have explored various approaches to measure the creativity of LLM. Broadly, these approaches fall into two distinct types. The first type involves adapting existing creativity assessment techniques from cognitive science for use with LLM. The second is to design a new approach specifically for measuring creativity in LLM.

For the first type, some researchers think that creative problem-solving is a crucial ability for LLM. For example,~\cite{stevenson2022putting} conducts a comparative analysis using the AUT, where both GPT-3 and human subjects were instructed to generate novel and useful responses. The responses were rated on a scale from 1 to 5 by two human judges, leading to the conclusion that humans outperformed GPT-3 in generating creative responses within the AUT. Building upon this,~\cite{SummersStay2023BrainstormTS} devised a series of prompts aimed at sifting through the 690 alternative uses responses previously generated, isolating those that were both original and practical. These prompts required the evaluation of the pros and cons associated with employing the objects in their new, unconventional roles. While GPT-3 demonstrated an ability to generate responses that were “surprisingly good,” it notably failed to discern and discard impractical alternative uses.~\cite{naeini2023large} introduces a novel dataset for evaluating creative problem-solving tasks by curating the problems and human performance results from the popular British quiz show Only Connect, an analogical proxy for RAT tests.~\cite{Cropley_2023} explore the creativity of LLM by using the DAT task on GPT-4 and GPT-3.5 compared to the human norms. In contrast to these approaches,~\cite{GesPushingGC} propose an innovative, interactive method enabling GPT-4 to autonomously refine and enhance the creativity of its outputs. They employ the AUT and the TTCT visual completion tasks to investigate the LLM's creativity. Similarly,~\cite{AI} also uses TTCT to evaluate the creative abilities of GPT-4.

For the second type,~\cite{wang2024ai} proves in theory that LLM can be as creative as humans under the condition that it can properly fit the data generated by human creators. Additionally, they also introduce two concepts of creativity in LLM: relative and statistical creativity. For relative creativity, where LLMs are deemed as creative as a hypothetical, yet realistic, human creator if it can produce works indistinguishable from that creator, as determined by an evaluator. Statistical creativity is a means for understanding whether and to what degree LLM achieves creativity by comparing it with existing human creators. Through the lens of statistical creativity. And ~\cite{lee2023mathematical} derive a mathematical characterization of the trade-off between hallucination and creativity in LLM by developing a rigorous mathematical framework for analyzing the hallucination phenomenon in LLMs.

Although researching creativity in LLMs become a trend, there is still a need for more theoretical studies and comprehensive benchmarks to deepen understanding of this area.

%% file: sec5.tex
\section{Harness LLM Hallucination For Creativity}~\label{sec5}

Given the recognition of LLM hallucinations as potential catalysts for creativity, a pivotal question arises: \textit{How can we effectively harness hallucinations for creative thought while also mitigating their risks?}
The aim is to enhance our understanding and utilization of these phenomena, thereby amplifying the creative potential of LLMs across various applications, without compromising on accuracy.
This requires methods that not only induce hallucinations but also critically evaluate them, transforming them into innovative sources.
In addressing this challenge, our research revolves around a comprehensive consideration of two critical dimensions: the divergent phase and the convergent phase. These phases align with the framework outlined in fundamental research~\cite{pressing1998psychological}, which posits creativity as an interplay of generative and evaluative processes, embracing both divergent and convergent thinking.
The divergent phase stimulates LLM hallucinations tailored to user needs, fostering creative thought, while the convergent phase assesses these to transform them into innovative, applicable ideas, and our survey will systematically explore studies aligned with these phases.

\subsection{Divergent Phase of LLM}
\begin{figure}[t]
	\centering
	\includegraphics[width=1\columnwidth]{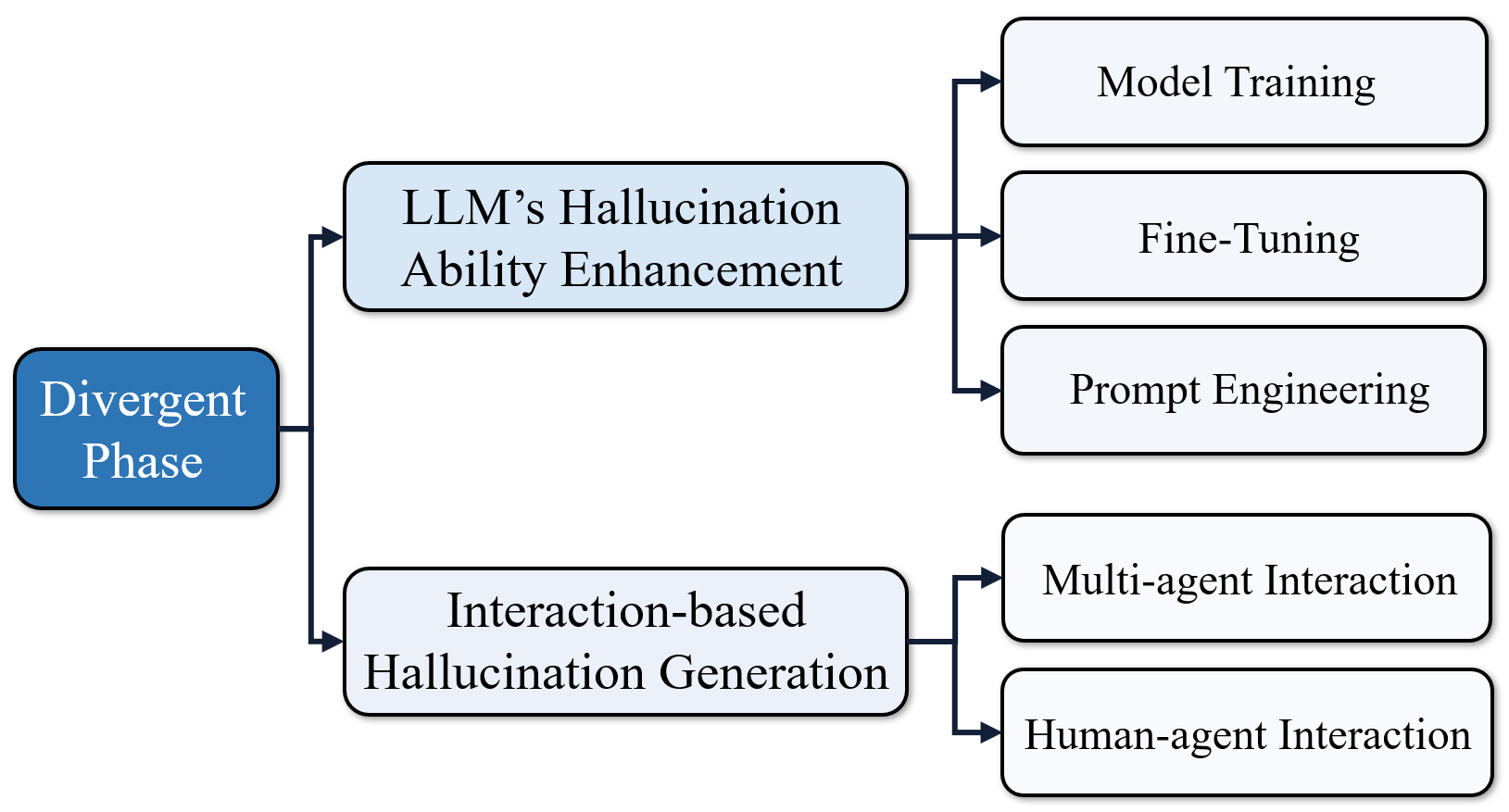}
	\caption{The divergent phase stimulates the hallucination of LLMs, fostering creative thought.}
	\label{FIG:DIV}
      \vspace{-10pt}
\end{figure}

In the divergent phase, the emphasis extends beyond the mere generation of relevant hallucinations. It encompasses fostering the LLM's capacity for hallucination with creativity. This phase benefits significantly from insights derived from existing literature, which suggest a multi-faceted approach to model optimization, as shown in Figure~\ref{FIG:DIV}.

For the hallucination ability enhancement of LLMs, Pressing's theory of improvisational creativity~\cite{pressing2007improvisation} suggests that improvisation is an acquired skill, requiring extensive training to reach professionals, which literature underscores the importance of training.
In concordance with the aforementioned views, recent studies~\cite{zhao2024assessing,tian2023macgyver} have proven the hallucination gap between different LLMs through extensive experiments, which points out the necessity of model training to improve the LLM's abilities of hallucination and creativity.
For instance, research~\cite{colin2016hierarchical} demonstrates how specific training techniques can enhance the model's ability to generate creative outputs. This research proposes that the cognitive process of creativity, especially the 'insight' phenomenon in problem-solving, aligns closely with the mechanisms of hierarchical reinforcement learning (HRL) in artificial systems. This work underscores the potential of HRL to mimic human-like creative processes, offering a novel perspective on the capabilities of LLM in replicating complex cognitive behaviors.
For prompt engineering, by setting specific prompts and contexts, LLMs will continuously learn and adapt in contexts to enhance their understanding and response capabilities for hallucination generation and complex creative tasks. This theory can be traced back to the classic literature of cognitive science. For example, giving a prompt like: "\textit{prompt: to come up with something clever, humorous, original, compelling, or interesting}" has been validated as effective for human hallucination in research~\cite{beaty2013first}.
In the prompt engineering research about LLM~\cite{hubert2023artificial,koivisto2023best}, it is also the most widely adopted strategy for divergent thinking, enabling LLMs to explore and respond to users' creative requirements more easily, thereby exhibiting a higher level of creativity in applications.

The integration of interaction-based hallucination generation methods also offers a promising direction to enrich the divergent phase. This concept introduces two distinct types of interactions: multi-agent interaction and human-agent interaction.
For the interaction among multiple agents, a dynamic that has been investigated in recent studies like~\cite{liang2023encouraging}, plays a crucial role. These multi-agent interactions encourage diverse and creative outputs from LLMs, as agents with varying perspectives and knowledge bases engage in collaborative or competitive dialogues.
For the human-agent interaction, an area receiving increasing attention as evidenced by~\cite{weber2023fool,rick2023supermind}, provides invaluable insights. This form of interaction allows for a more grounded and realistic feedback loop, where human creativity can inspire and be inspired by the LLM's responses, leading to more relatable and applicable hallucinations.
In summary, these two types of interactions underpin the diversification of hallucination generation in LLMs, each contributing to a more dynamic, context-aware, and creatively stimulating phase.

As examined in \cite{koivisto2023best}, the divergent thinking ability of LLM is still far from human, which requires the improving of LLM divergent thinking abilities.


\subsection{Convergent Phase of LLM}

The convergent phase, in stark contrast to the divergent phase, is dedicated to refining hallucinations into valuable creative contributions. This phase is characterized by a meticulous process of evaluating hallucinations, employing a systematic approach to identify, categorize, and select those that are most constructive and aligned with user needs.

\begin{figure}[t]
	\centering
	\includegraphics[width=1\columnwidth]{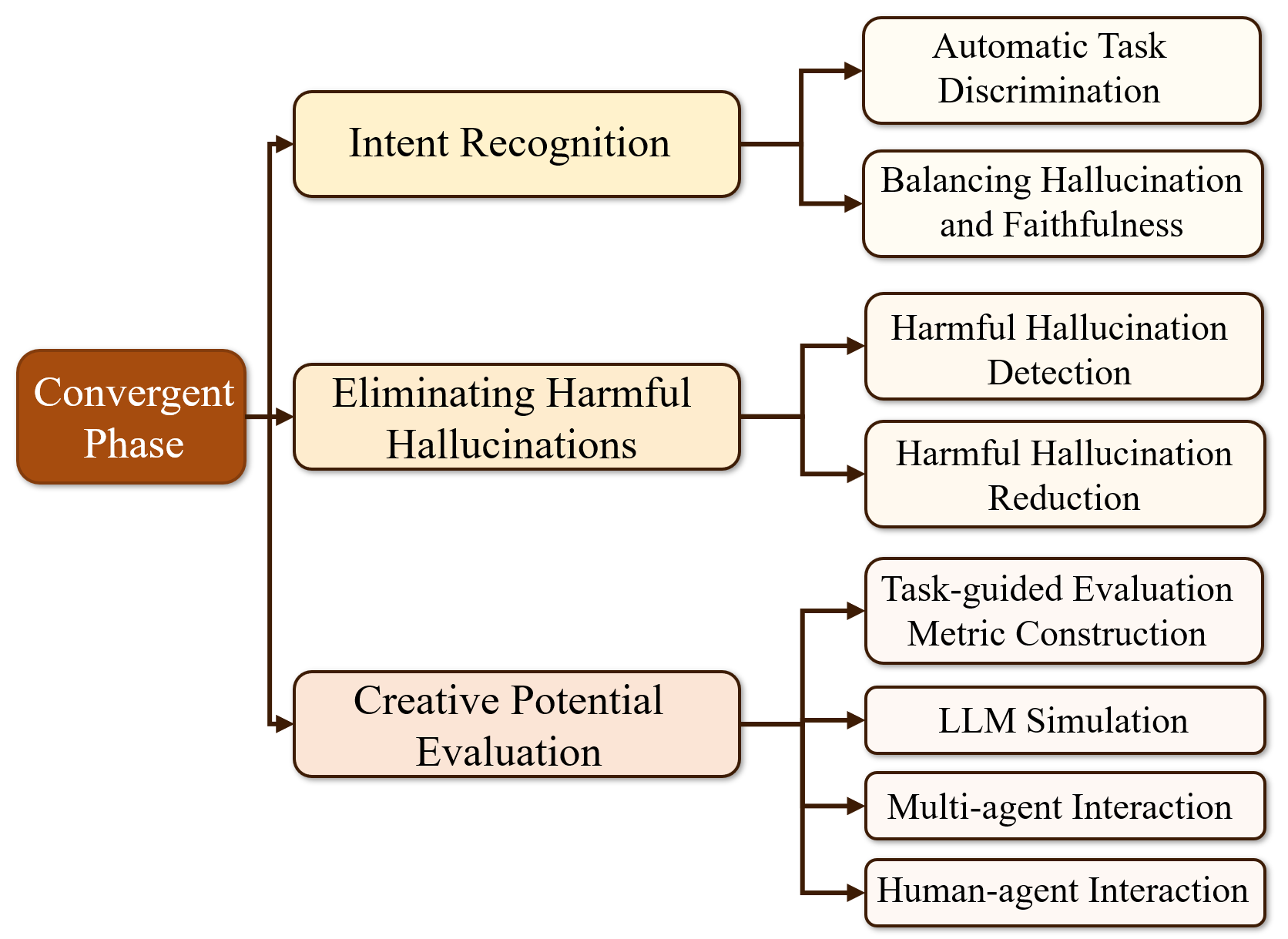}
	\caption{The convergent phase refines hallucinations into valuable creative contributions.}
	\label{FIG:CONV}
        \vspace{-10pt}
\end{figure}

Different from existing hallucination detection and reduction methods as discussed in Sections~\ref{hallucination_reduction} and~\ref{hallucination_detection}, which directly identify and eliminate hallucinations, the convergent phase involves discerning whether hallucinations are needed based on the context, eliminating harmful hallucinations, and identifying valuable ones, raising higher demands for LLM convergent thinking abilities, as shown in Figure~\ref{FIG:CONV}.

To determine whether hallucinations are necessary for applications, it's crucial to optimize the intent recognition capabilities of LLMs.
For instance, the 4C creativity theory discussed in research~\cite{tian2023macgyver} explores four scenarios requiring creativity, optimizing Little-C tasks for LLMs' creativity. Automating the discrimination of these task scenarios remains an area of urgent research.
Moreover, balancing hallucination and faithfulness in LLMs based on the demand, as mentioned in research~\cite{zhang2023user}, involves fine-tuning and knowledge fusion, but this area is still in its infancy, and lacking a systematic approach.

To discern harmful hallucinations, a detailed and nuanced classification system for hallucinations is crucial. Existing research, as extensively discussed in Section~\ref{hallucination_detection}, has developed a taxonomy for hallucinations predominantly under the assumption that they are detrimental and need to be mitigated. However, this approach overlooks the potential value of certain hallucinations. The current classification lacks in annotating and recognizing beneficial hallucinations, thereby limiting the LLMs' ability to differentiate between harmful and potentially valuable hallucinations.

To evaluate the creative potential hidden within the sea of hallucinations produced by LLMs, it's crucial to establish metrics and design methods for effectively assessing LLMs' creative output.
In terms of evaluation metrics, as discussed in Section~\ref{creativity_evaluate}, the evaluation of creativity within these hallucinations can leverage existing metrics from cognitive science and those specifically developed for LLMs. The choice of metrics should align with the specific applications. For instance, artistic creation often emphasizes fluency, flexibility, originality, and elaboration~\cite{silvia2008assessing}, whereas scientific innovation is more focused on problem-solving abilities~\cite{tian2023macgyver}.
In terms of evaluation methods, those mentioned in Section~\ref{creativity_evaluate} for assessing creativity are still applicable to hallucinations. However, these studies often rely on manual assessment by data annotators or experts. Automating the assessment of creativity within hallucinations is key to transitioning from theory to practical applications. From the model's ability perspective, recent studies, such as~\cite{tian2023macgyver,zhao2024assessing}, propose self-reflection for automatic content evaluation and correction but highly rely on the model's reasoning abilities, which requires effective training methods or prompt strategies in future research.
From the interaction perspective, studies like ChatEval~\cite{chan2023chateval} attempt to use multi-agent approaches to assess whether generated content meets requirements. Research~\cite{zhang2023creative} evaluates the creative value behind LLM actions from a simulation perspective. Research from the human-agent perspective~\cite{rick2023supermind}, assessing user interactions with LLM-generated content, offers a way to evaluate creativity in hallucinations, moving away from traditional manual annotation.

In conclusion, while many studies have explored this direction, research in this field is still in its early stages. A more comprehensive evaluation metric system, richer datasets, and stronger models and framework designs are urgently needed for further exploration.

%% file: sec6.tex
\section{Conclusion and Future Work}~\label{sec6}
\vspace{-15pt}

\subsection{Conclusion}
This paper revisits the hallucinations in LLMs, advocating for a paradigm shift where they are not solely viewed as negative phenomena, but also as potential catalysts for creativity. It establishes what hallucinations entail within LLMs and reviews research focused on their detection and reduction. Challenging the view of hallucinations as entirely negative, the study delves into their potential to spur creativity. Examining historical cases and current research highlights the intricate interplay between hallucination and creativity in LLMs. The discussion follows the cognitive science framework of divergent and convergent thinking, providing a comprehensive review of studies on harnessing creative hallucinations. This work not only aligns with cognitive science principles but also opens avenues for applications and future research in leveraging hallucinations for creative purposes in LLMs.


\subsection{Future Work}
Future research in the field of hallucinations and creativity within LLMs can be explored in several key areas:

\textbf{Deeper Theoretical Exploration:} Although cognitive science has extensively studied the interplay between hallucinations and creativity in humans as illustrated in Section~\ref{sec3}, there's a need for more theoretical exploration within the context of LLMs. Future research should aim to deepen our understanding of this relationship in LLMs, focusing on constructing a theoretical basis that connects cognitive theories with the operational mechanisms of LLMs.

\textbf{Richer Datasets and Benchmarks:} As discussed in Section~\ref{sec5}, there is a notable lack of available benchmarks and datasets for the study of hallucination and creativity. Future research urgently requires the development of richer datasets to facilitate research and experimentation in this domain. Additionally, the establishment of more comprehensive benchmarks is crucial for a deeper evaluation of the abilities of LLMs in manifesting creative outputs from hallucinations.

\textbf{Optimization of Method Designs:} As discussed in Section~\ref{sec5}, leveraging hallucinations for creativity relies on the LLMs' intrinsic reasoning capabilities to discern user intent, distinguish harmful hallucinations, and evaluate the creativity within hallucinations. Future research should thus focus on optimizing LLMs' abilities in these areas through advancements in training strategy and framework design, ensuring effective utilization of hallucinations for creative output.

\textbf{Transformation of Hallucinations to Creativity in Multimodal Scenarios:} As explored in Section~\ref{sec5}, the value of multimodal information during the divergent and convergent phases of creativity highlights a nascent yet challenging area of research. Future investigations are needed to tackle how to assess the creative value of hallucinations across different modalities and use multimodal information for creativity.

\textbf{Exploration of More Application Scenarios:} In Sections~\ref{sec4} and~\ref{sec5}, we examined various application fields where users' demands for creativity differ, leading to diverse standards for assessing the value of creativity within hallucinations. This underscores the necessity for future investigations to explore a broader range of application scenarios. Such exploration is essential to achieve a more comprehensive understanding and utilization of the interplay between hallucination and creativity across different contexts.

Ultimately, these directions for future research endeavor to enrich the comprehension and application of hallucinations in LLMs, contributing innovative theories and methodologies and paving the way for practical applications.




